\documentclass[10pt]{article}

\usepackage{times}
\usepackage{epsfig}
\usepackage{graphicx}
\usepackage{amsmath}
\usepackage{amssymb}
\usepackage{algorithm,algorithmic}
\usepackage{bm,upgreek}
\newcommand{\Rho}{\mathrm{P}}
\newcommand{\argmax}{\mathrm{argmax}}

\begin{document}

\title{Maximum A Posteriori Estimation of Distances Between Deep Features in Still-to-Video Face Recognition}

\author{Andrey V. Savchenko\\
National Research University Higher School of Economics\\
Laboratory of Algorithms and Technologies for Network Analysis, \\36 Rodionova St., Nizhny Novgorod, Russia\\
{\tt\small avsavchenko@hse.ru}
\and
Natalya S. Belova\\
National Research University Higher School of Economics\\
20 Myasnitskaya St., Moscow, Russia\\
{\tt\small nbelova@hse.ru}
}

\maketitle
\begin{abstract}
The paper deals with the still-to-video face recognition for the small sample size problem based on computation of distances between high-dimensional deep bottleneck features. We present the novel statistical recognition method, in which the still-to-video recognition task is casted into Maximum A Posteriori estimation. In this method we maximize the joint probabilistic density of the distances to all reference still images. It is shown that this likelihood can be estimated with the known asymptotically normal distribution of the Kullback-Leibler discriminations between nonnegative features. The experimental study with the LFW (Labeled Faces in the Wild), YTF (YouTube Faces) and IJB-A (IARPA Janus Benchmark A) datasets has been provided. We demonstrated, that the proposed approach can be applied with the state-of-the-art deep features and dissimilarity measures. Our algorithm achieves 3-5\% higher accuracy when compared with conventional aggregation of decisions obtained for all frames.
\end{abstract}


\section*{Abbreviations}
\begin{itemize}
\item CNN - Convolution Neural Network
\item i.i.d. - independent identically distributed
\item IJB-A - IARPA Janus Benchmark A
\item KL - Kullback-Leibler divergence
\item LFW - Labeled Faces in the Wild
\item MAP - Maximum A Posteriori
\item ML - Maximum Likelihood
\item NN - Nearest Neighbor 
\item YTF - YouTube Faces
\end{itemize}

\section{Introduction}
\label{sec:1}
The usage of deep learning technologies~\cite{lecun2015deep} instead of traditional computer vision methods~\cite{prince2012computer} has recently made it possible to achieve a near human-level performance in various face recognition tasks, such as verification~\cite{taigman2014deepface} or identification~\cite{liu2015targeting}. Moreover, contemporary techniques can deal even with well-known crucial issues appeared in practical applications of face recognition, e.g., unconstrained environment (various illumination and pose, partial occlusion)~\cite{learned2016labeled}, or the small sample size problem~\cite{savchenko2016search}, when usually only single facial image per person is available~\cite{savchenko2015statistical,tan2006face}. The latter problem is solved using transfer learning methods~\cite{cao2013practical}, in which large external datasets of celebrities are used to train deep convolutional neural network (CNN)~\cite{parkhi2015deep,wu2016light}. This CNN is further applied to extract features of the training images from the limited sample of subjects of interest using the outputs of a bottleneck layer~\cite{savchenko2017deep}. Such transfer learning technique makes it possible to create a classifier that ideally performs nearly as well as if rich dataset of photos of these individuals were present~\cite{cao2013practical}. 

Though the algorithms that do recognition from a single test photo have its own practical applications, e.g., search for a person in a social network, one of the most significant use-cases is provided by surveillance and authentication systems. Hence, all the more attention is nowadays paid to video face recognition~\cite{cui2013fusing}, in which a set or a sequence of observed images of the same individual are available for decision-making~\cite{cevikalp2010face,mian2013image,wolf2003kernel}. Despite the large number of recent papers devoted to video face recognition~\cite{huang2013coupling,liu2014toward,zhu2015still,yang2016neural}, this task still remains very challenging. For example, most of the known algorithms suffered from heavy off-line training load~\cite{liu2014toward}, and they were experimentally studied only with databases, in which the still images are gathered under controlled environment. In this paper we proposed the novel still-to-video face recognition method suitable for application with the small sample of still photos per person, which is based on the probabilistic interpretation~\cite{savchenko2016search} without learning temporal coherence~\cite{liu2015targeting}. At first, the nearest neighbor (NN) reference images are examined for each video frame. Next, computed distances to all still images are used to weight the recognition results for each frame based on the estimation of their reliability. The more is the likelihood of the computed vector of distances for particular individual, the more is the weight corresponding to this frame and this subject. The likelihoods (joint probabilistic densities) are computed using the idea of the maximum-likelihood approximate NN algorithm~\cite{savchenko2017maximum} by assuming that the Kullback-Leibler (KL) divergence~\cite{kullback1997information} is used to compare distances. However, we demonstrate that our approach can be successfully applied with much more widely used dissimilarity measures.

The rest of the paper is organized as follows. Section~\ref{sec:2} contains the literature survey of related papers. In Section~\ref{sec:3} we present a simple statistical formulation of the still-to-video recognition task~\cite{shakhnarovich2002face} using the KL minimum discrimination principle. In Section~\ref{sec:4} we propose the novel approach, in which the maximum a-posteriori (MAP) rule is regularized using the computation of the joint probability densities of distances based on the asymptotic properties of the KL divergence. In Section~\ref{sec:5} we present the complete algorithm, in which the initial assumptions are facilitated and the computational complexity is taken into account. Section~\ref{sec:6} presents the experimental results in recognition of videos from either IJB-A (IARPA Janus Benchmark A)~\cite{klare2015pushing} or YouTube Faces (YTF) datasets~\cite{wolf2011face} with the still images from the Labeled Faces in the Wild (LFW) dataset~\cite{learned2016labeled} using such deep CNNs as VGGNet~\cite{parkhi2015deep} and Lightened CNN~\cite{wu2016light}. Finally, concluding comments are given in Section~\ref{sec:7}.

\section{Related works}
\label{sec:2}
Compared to still-to-still recognition, the requirement of real-time processing in video face recognition prevents an implementation of too sophisticated recognition algorithms~\cite{savchenko2012adaptive}, especially, when the number of individuals is rather large~\cite{savchenko2017maximum}. However, more information of the individuals can be exploited from several video frames~\cite{yang2016neural}, and the recognition results can be generally improved when using multiple images or video sequences~\cite{zhang2006weighted}. Most of the papers devoted to person identification or retrieval in surveillance videos~\cite{cevikalp2010face,liu2003video,mian2013image,wolf2011face,yang2016neural} address the video-to-video face recognition problem, in which query videos are matched against a set of reference video sets~\cite{liu2014toward}. However, real-world applications usually involve the still-to-video face recognition scenario, when only very few still images per person are enrolled into the training set while the input sequence of video frames is captured~\cite{huang2013coupling,zhu2015still}. This problem is very difficult, because the testing videos have generally low quality and are captured under poor illumination and pose~\cite{huang2012benchmarking}. 

There are two types of still-to-video recognition methods~\cite{liu2014toward}. First group of methods exploits the temporal information in a video sequence. Liu and Chen used the hidden Markov models in recognition of video frames~\cite{liu2003video}. The paper~\cite{zhou2002face} describes successful application of a state space model~\cite{prince2012computer} parametrized by a tracking state vector and a recognizing identity variable, simultaneously characterizing the kinematics and identity of humans. All such methods require large datasets of video clips to train the extraction of dynamic information. Moreover, their performance relied on strong temporal coherence in training and testing environment~\cite{liu2014toward}. Hence, another group of methods with accumulation of multiple observations of the same individual has recently been studied. These methods are usually based on fusion of evidence from multiple measurements~\cite{kittler1998combining}. For example, the authors of the paper~\cite{zhang2006weighted} extended the probabilistic appearance-based face recognition approach to work with multiple images and video sequences. Shakhnarovich et al~\cite{shakhnarovich2002face} assumed that the multiple frames for each human subject followed the normal distribution, and they used the KL divergence~\cite{kullback1997information} to measure the distance between distributions. Several important works on still-to-video face recognition adopted a metric learning framework. For example, Zhu et al~\cite{zhu2013point} extended the well-known Mahalonbis metric learning to point-to-set and set-to-set distance learning. Huang et al~\cite{huang2014learning} applied Euclidean-to-Riemannian metric learning framework to image-to-set object classification and still-to-video face recognition tasks. 

Nowadays, it is more typical to map each face image into a feature vector using a deep CNN. Next, a naive approach would be representing a video face as a set of frame-level face features from the CNN~\cite{schroff2015facenet}. Such a representation obviously maintains all the information across all frames. To speed-up the recognition process, the features of all video frames are aggregated into a single vector (video face representation) using such pooling strategies as average and max pooling~\cite{chen2015end,li2014eigen}. The Eigen-PEP representations~\cite{li2014eigen} integrates the visual information from all relevant video frames using part-based average pooling through the probabilistic elastic part model. Then the intermediate representation is compressed through principal component analysis, and only a number of principal eigen dimensions are kept. Yang et al~\cite{yang2016neural} proposed the learning of the frames weights in the aggregation module, which consists of two attention blocks driven by a memory storing all the extracted features. Canziani and Culurciello proposed the CortexNet neural architecture~\cite{yang2017cortexnet} in order to obtain a robust and stable representation of temporal visual inputs. In the paper~\cite{wolf2003kernel} two video sequences representing head motion and facial expressions are compared using a new positive definite kernel based on the concept of principal angles between two linear subspaces. Most of described methods learned the relationship between the still images and video frames but did not directly handle bad quality frames, which very likely make the recognition perform badly. To deal with this problem, another kind of methods such as~\cite{wong2011patch} was proposed, which first select the best quality frames, perform facial alignments and then integrate the recognition results of selected frames~\cite{huang2013coupling}. Thus, one can notice from this brief survey, that the methods without learning temporal coherence are more widely used nowadays. Hence, in our paper we decided to focus on application of statistical recognition of individual frames~\cite{shakhnarovich2002face} described by the modern deep CNN features~\cite{lecun2015deep}.

\section{Statistical still-to-video face recognition}
\label{sec:3}

The task of closed-set still-to-video face identification is to assign an observed sequence of $T$ video frames to one of $C>1$ classes (identities). The classes are specified by the training set of $R\ge C$ still images. We consider the supervised learning case, when the class label (subject id) $c(r) \in \{1,...,C\}$ of the $r$th photo is known. For simplicity, we assume that only one image per person is available ($C=R,c(r)=r$), and all frames contain facial region of only one identity, i.e., the faces are detected, and the whole video clip is clustered to extract sequential frames containing the same subject. At first, each image should be described with a feature vector using the preliminarily trained deep CNN, as it was described in introduction. The outputs at the last layer of this CNN for all $t$ frames and each $r$-th reference image are used as the $D$-dimensional feature vectors $\mathbf{x}(t)=[x_1(t),...,x_D(t)]$ and $\mathbf{x}_r=[x_{r;1},...,x_{r;D}]$, respectively. 

Let us assume that the normalized feature vector $\mathbf{x}(t)$ is the estimate of multinomial distribution of (hypothetical) random variable $X(t),t \in \{1,...,T\}$. We also assume that each $r$th instance represents the probabilistic distribution of the random variable $X_r,r \in \{1,...,R\}$. The video face recognition task is reduced to a problem of statistical testing of simple hypothesis $W_r, r \in \{1,...,R\}$ for all frames. Statistically optimal decision is defined as follows:

\begin{equation}
\underset{r \in \{1,...,R\}}\max P(W_r|\mathbf{x}(1),...,\mathbf{x}(T)),
\end{equation}
where the posterior probability is estimated using the Bayes rule:

\begin{equation}
P(W_r|\mathbf{x}(1),...,\mathbf{x}(T))=\frac{f(\mathbf{x}(1),...,\mathbf{x}(T)|W_r)}{\sum\limits_{i=1}^{R} {f(\mathbf{x}(1),...,\mathbf{x}(T)|W_i)}},
\end{equation}

In this paper we focus on the case of full prior uncertainty, hence, the prior probabilities of observing each subject here are equal, so the rule (1) represents the maximum likelihood (ML) criterion. Finally, it is necessary to perform aggregation of the frame recognition results~\cite{savchenko2012adaptive}. Kittler et al~\cite{kittler1998combining} presented a statistical interpretation of a number of common methods for cross-modal fusion, such as the product, sum, maximum, and majority vote rules, which are also appropriate for late integration over a set of observations from a single modality. For example, it can be shown that under the assumption that the frames in the video sequence correspond to independent identically distributed (i.i.d) random variables, the likelihood is estimated using the product rule~\cite{shakhnarovich2002face}

\begin{equation}
f(\mathbf{x}(1),...,\mathbf{x}(T)|W_r)=\prod\limits_{t=1}^{T} {\prod\limits_{d=1}^{D} {(x_{r;d})^{n \cdot x_d(t)}}},
\end{equation}
where $n$ is the sample size used to estimate the distribution $\mathbf{x}(t)$. In practice, this parameter can be chosen proportional to the total number of pixels in the input image. By substituting (3) in (2) and dividing its numerator and denominator to $\prod\limits_{t=1}^{T} {\prod\limits_{d=1}^{D} {(x_d(t))^{n \cdot x_d(t)}}}$, one can obtain the following estimate of the posterior probability:

\begin{equation}
P(W_r|\mathbf{x}(1),...,\mathbf{x}(T))= \frac{\exp(-n \sum\limits_{t=1}^{T} {\sum\limits_{d=1}^{D} {x_d(t) \ln(\frac{x_d(t)}{x_{r;d}})}})}{\sum\limits_{i=1}^{R} {\exp(-n \sum\limits_{t=1}^{T} {\sum\limits_{d=1}^{D} {x_d(t) \ln(\frac{x_d(t)}{x_{i;d}})}})}},
\end{equation}
or
\begin{equation}
P(W_r|\mathbf{x}(1),...,\mathbf{x}(T))= \frac{\exp\left(-n \sum\limits_{t=1}^{T} {I(\mathbf{x}(t):\mathbf{x}_r)}\right)}{\sum\limits_{i=1}^{R} {\exp\left(-n \sum\limits_{t=1}^{T} {I(\mathbf{x}(t):\mathbf{x}_i)}\right)}},
\end{equation}
where $I(\mathbf{x}(t):\mathbf{x}_r))$ is the KL divergence between feature vectors $\mathbf{x}(t)$ and $\mathbf{x}_r$.

Hence, the ML criterion (1) can be written in the simplified form:

\begin{equation}
\underset{r \in \{1,...,R\}}{\min} \sum\limits_{t=1}^{T} {I(\mathbf{x}(t):\mathbf{x}_r)}.
\end{equation}

Thus, the ML decision (1) of the still-to-video face recognition task~\cite{shakhnarovich2002face} and assumptions about i.i.d. frames is equivalent to the KL minimum information discrimination principle~\cite{kullback1997information}.

\section{Maximum a posteriori estimation of distances between deep bottleneck features}
\label{sec:4}

In this paper we use the idea of the maximal-likelihood approximate NN method from the paper~\cite{savchenko2017deep,savchenko2017maximum}. Namely, we exploit the known property of the KL divergence between two densities, which can be considered as the information for discrimination in favor of the first density against the second one~\cite{kullback1997information}. Hence, another recognition criterion is used, in which the maximum posterior probability based on the joint distribution of all $R$-dimensional random vectors of \textit{distances} $\mathbf{I}(t)=[I(\mathbf{x}(t):\mathbf{x}_1),..., I(\mathbf{x}(t):\mathbf{x}_R)]$ is obtained:

\begin{equation}
\underset{r \in \{1,...,R\}}\max f(W_r|\mathbf{I}(1),...,\mathbf{I}(T)).
\end{equation}

To deal with this conditional density of distances let us firstly consider much more simple marginal conditional distribution $f(\mathbf{I}(t)|W_r), t \in \{1,...,T\}$. By using a natural assumption about independence of the still photos of all identities from the training set, we estimate the joint density of distances between reference images and the $t$th frame $f(\mathbf{I}(t)|W_r)$ as follows:

\begin{equation}
f(\mathbf{I}(t)|W_r)=f(I(\mathbf{x}(t):\mathbf{x}_r)|W_r) \cdot \prod\limits_{i=1,i \neq r}^{R} {f(I(\mathbf{x}(t):\mathbf{x}_i)|W_r)}.
\end{equation}

At first, let us consider each term in the second multiplier. We propose to estimate the conditional density function $f(I(\mathbf{x}(t):\mathbf{x}_i)|W_r)$ of the distance between the observed object from the $r$th class and the $i$th instance using the known asymptotic properties of the KL divergence. It is known~\cite{kullback1997information} that if $i \neq r$, then $2n$-times KL divergence $I(\mathbf{x}(t):\mathbf{x}_r)$ is asymptotically distributed as the non-central chi-squared with $D-1$ degrees of freedom and the non-centrality parameter $2n \cdot I(\mathbf{x}_r:\mathbf{x}_i)$. This asymptotic distribution is also known to be achieved for such probabilistic dissimilarities as the chi-squared distance, the Jenson-Shannon divergence, etc.~\cite{savchenko2017deep,kullback1997information}. As the number of features $D$ is usually large, it is possible to approximate the non-central chi-squared distribution with the Gaussian distribution~\cite{savchenko2017maximum}. Hence, we use the asymptotically normal distribution of the distance between the facial image from the $r$th class and the $i$th reference photo:
\begin{equation}
\mathcal{N}\left(I(\mathbf{x}_r:\mathbf{x}_i)+\frac{D-1}{n},\frac{4n \cdot I(\mathbf{x}_r:\mathbf{x}_i)+(D-1)}{2n^2}\right)
\end{equation}

Thus, the conditional distributions in the second multiplier in (8) are estimated as follows:
\begin{equation}
f(I(\mathbf{x}(t):\mathbf{x}_i)|W_r) = \exp( -n \cdot \phi_{r;i} (\mathbf{x}(t))),
\end{equation}
where we denote $\phi_{r;i} (\mathbf{x}(t))$ for 
\begin{multline}
\phi_{r;i} (\mathbf{x}(t)) = \frac{1}{2n}\ln \left (4I(\mathbf{x}_r:\mathbf{x}_i)+\frac{\pi+D-1}{n}\right)+\\+\frac{(I(\mathbf{x}(t):\mathbf{x}_i)-I(\mathbf{x}_r:\mathbf{x}_i)-(D-1)/n)^2}{4I(\mathbf{x}_r:\mathbf{x}_i)+(D-1)/n}.
\end{multline}

On the one hand, the number of outputs (features) in the last layer in modern DNNs is very high ($D \gg 1$)~\cite{parkhi2015deep,wu2016light}. On the other hand, the feature vector dimensionality $D$ should be much less than the number of samples $n$ to estimate the conditional densities (3) reliably ($D \ll n$). Hence, it is possible to approximately rewrite eq. (11)~\cite{savchenko2017maximum}:

\begin{equation}
\phi_{r;i} (\mathbf{x}(t))\approx \frac{(I(\mathbf{x}(t):\mathbf{x}_i)-I(\mathbf{x}_r:\mathbf{x}_i))^2}{4I(\mathbf{x}_r:\mathbf{x}_i)}.
\end{equation}

The first multiplier in (8) cannot be computed similarly, because, if $i=r$, then $I(\mathbf{x}_r:\mathbf{x}_i)=0$, and asymptotic distribution (9) does not hold in practice. However, in such a case this conditional density can be estimated using eq. (3). Similarly to (4), (5) one can show that 

\begin{equation}
f(I(\mathbf{x}(t):\mathbf{x}_r)|W_r)) \propto \exp\left(-n I(\mathbf{x}(t):\mathbf{x}_r)\right).
\end{equation}

By combining expressions (10), (12), (13), the marginal distribution (8) is written using softmax operation as follows
 
\begin{equation}
f(\mathbf{I}(t)|W_r)=\frac{\exp\left(-n \cdot (I(\mathbf{x}(t):\mathbf{x}_r)+\sum\limits_{i=1}^{R} {\phi_{r;i} (\mathbf{x}(t))})\right)}{\sum\limits_{j=1}^{R}{\exp\left(-n \cdot (I(\mathbf{x}(t):\mathbf{x}_j)+\sum\limits_{i=1}^{R} {\phi_{j;i} (\mathbf{x}(t))})\right)}}.
\end{equation}

After that it is possible to apply some known rules~\cite{kittler1998combining} of combining statistical classifiers (14). For example, in case of the i.i.d. frames assumptions the criterion (7) is equivalent to rather simple video face recognition rule
\begin{equation}
\underset{r \in \{1,...,R\}}\min \sum\limits_{t=1}^{T} {\left(I(\mathbf{x}(t):\mathbf{x}_r)+\sum\limits_{i=1}^{R} {\phi_{r;i} (\mathbf{x}(t))}\right)},
\end{equation}
which can be viewed as an extension of conventional statistical classification (6) using the new regularization term $\sum\limits_{i=1}^{R} {\phi_{r;i} (\mathbf{x}(t))}$. However, Kittler et al~\cite{kittler1998combining} argued, that the sum rule is much more robust to the errors in unknown density estimation. In our task this rule can be formulated in the MAP way:
\begin{equation}
\underset{r \in \{1,...,R\}}\max \sum\limits_{t=1}^{T} {f(W_r|\mathbf{I}(t))},
\end{equation}
where posterior probability $f(W_r|\mathbf{I}(t))$ is defined using the Bayes rule (2) for the proposed ML of distances estimate (14).

\section{Proposed algorithm}
\label{sec:5}

Though the previous sections contain many unrealistic assumptions about independence of sequential frames and representation of feature vectors as the estimates of the distributions of hypothetical random variables, the resulted criteria (15), (16) looks rather straightforward. In this section we demonstrate the possibility to implement this approach in more realistic scenarios. At first, it is necessary to emphasize, that the criteria discussed in the previous section contain only computations of the KL divergence between features of all frames and every instances from the database (12). Thus, it is possible to replace the KL divergence $I(\mathbf{x}(t):\mathbf{x}_r)$ in (12), (13) with an arbitrary dissimilarity measure $\rho(\mathbf{x}(t),\mathbf{x}_r)$ between deep bottleneck features $\mathbf{x}(t)$ and $\mathbf{x}_r$. For instance, the KL minimum information discrimination criterion (6) is equivalent to the NN rule with the simple accumulation~\cite{kittler1998combining}:

\begin{equation}
\underset{r \in \{1,...,R\}}{\min} \sum\limits_{t=1}^{T} {\rho(\mathbf{x}(t),\mathbf{x}_r)}.
\end{equation}

The sum rule~\cite{kittler1998combining} implements the MAP estimate of the the class, which corresponds to the maximal average posterior probability~\cite{savchenko2012adaptive}:

\begin{equation}
\underset{c \in \{1,...,C\}}{\max} \sum\limits_{t=1}^{T} {\frac{\exp(-n\cdot \rho_c(\mathbf{x}(t))}{\sum\limits_{i=1}^{C}{\exp(-n\cdot \rho_i(\mathbf{x}(t))}}},
\end{equation}

It is important to emphasize that our conclusion about asymptotical distribution (10) of the KL divergence is in agreement with the well-known assumption about Gaussian distribution of dissimilarity measures between high-dimensional feature vectors, which is supported by many experiments~\cite{burghouts2008distribution}.

Secondly, though the small sample size problem is usually appeared in practical applications of video face recognition, it is not typical to have only one instance per class in the available training set. Hence, we propose to extend criterion (15) to the case of $C\le R$:
\begin{equation}
\underset{c \in \{1,...,C\}}\max \sum\limits_{t=1}^{T} {\frac{\exp\left(-n \cdot (\rho_c(\mathbf{x}(t))+\frac{\lambda}{C} \sum\limits_{i=1}^{C} {\frac{(\rho_i(\mathbf{x}(t))-\rho_{c;i})^2}{\rho_{c;i}}})\right)}{\sum\limits_{j=1}^{C}{\exp\left(-n \cdot (\rho_j(\mathbf{x}(t))+\frac{\lambda}{C} \sum\limits_{i=1}^{C} {\frac{(\rho_i(\mathbf{x}(t))-\rho_{j;i})^2}{\rho_{j;i}}})\right)}}}.
\end{equation}

Here we added the smoothing parameter $\lambda>0$ to tune the influence of our regularization. The distance between feature vector $\mathbf{x}(t)$ and the $c$th identity in this criterion can be defined using the idea of the single-linkage clustering:
\begin{equation}
\rho_c(\mathbf{x}(t))=\underset{r \in \{1,...,R\},c(r)=c}\min \rho(\mathbf{x}(t),\mathbf{x}_r).
\end{equation}

The distance between images from $c$th and $i$th identities can be computed as the average distance between instances from these classes:
\begin{equation}
\rho_{c;i}=\frac{1}{R_cR_i} \sum\limits_{r=1}^{R} {\sum\limits_{r_i=1}^{R} {\delta(c-c(r))\delta(i-c(r_i))\rho(\mathbf{x}_r,\mathbf{x}_{r_i})}},
\end{equation}
where $\delta(c)$ is the discrete delta function (indicator) and $R_c=\sum\limits_{r=1}^{R} {\delta(c-c(r))}$ is the total number of still photos of the $c$th subject. 

Finally, it is necessary to notice that the run-time complexity of the proposed criterion (19) $O(T(R+C^2
))$ is more than $C$-times higher than the complexity of the baseline ML rule (17). Hence, our approach cannot be applied in practical tasks when the number of classes is rather large~\cite{savchenko2017maximum}. To speed-up the decision process, we propose to modify criterion (19) by examining only $M \ll C$ candidate classes. Namely, the sum of distances between all frames and each class are computed identically to the conventional approach (17):
\begin{equation}
\rho_c(\mathbf{x}(1),...,\mathbf{x}(T))=\frac{1}{T}\sum\limits_{t=1}^{T} {\rho_c(\mathbf{x}(t))},
\end{equation}
and $M$ candidate classes $\{c_1,...,c_M\}$ with the lowest distances (18) are chosen. This $M$th element search (selection algorithm) is known to have linear average complexity: $O(C)$. In the final decision only these $M$ candidates are checked, so the criterion (19) is modified as follows:

\begin{equation}
\underset{c \in \{c_1,...,c_M\}}\max \sum\limits_{t=1}^{T} {\frac{\exp\left(-n \cdot (\rho_c(\mathbf{x}(t))+\frac{\lambda}{C} \sum\limits_{i=1}^{C} {\frac{(\rho_i(\mathbf{x}(t))-\rho_{c;i})^2}{\rho_{c;i}}})\right)}{\sum\limits_{m=1}^{M}{\exp\left(-n \cdot (\rho_{c_m}(\mathbf{x}(t))+\frac{\lambda}{C} \sum\limits_{i=1}^{C} {\frac{(\rho_i(\mathbf{x}(t))-\rho_{c_m;i})^2}{\rho_{c_m;i}}})\right)}}}.
\end{equation}

\begin{algorithm}
\caption{Maximum likelihood of distances in still-to-video recognition} 
\label{algorithm1}
\begin{algorithmic}[1]
\REQUIRE observed sequence of $T$ frames, database of facial feature vectors $\{\mathbf{x}_r\}$
\ENSURE subject id (class label) corresponded to the objective function (23)
\STATE Initialize arrays of distances $\Rho[c]:=0, \rho[c][t]:=0, c \in \{ 1,...,C\}, t \in \{ 1,...,T\}$
\FOR{each frame $t \in \{1,...,T\}$}
\STATE Detect facial region in the $t$th frame and extract deep features $\mathbf{x}(t)$ from bottleneck layer 
\FOR{each class label $c \in \{1,...,C\}$}
\STATE Assign $\rho[c][t]=\rho_c(\mathbf{x}(t))$ (20)
\STATE Assign $\Rho[c]:=\Rho[c]+\rho[c][t]$ 
\ENDFOR
\ENDFOR
\STATE Obtain labels $\{c_1,...,c_M\}$ using a selection algorithm for finding $M$ smallest numbers in the array $\{\Rho[c] | c \in \{ 1,...,C\}\}$
\STATE Initialize posterior probabilities $L[m]:=0, m \in \{ 1,...,M\}$
\FOR{each frame $t \in \{1,...,T\}$}
\STATE Assign $s:=0$, initialize posterior probabilities $L_t[m]:=0, m \in \{ 1,...,M\}$
\FOR{each top class index $m \in \{1,...,M\}$}
\STATE Assign $\Omega:=0$
\FOR{each class label $i \in \{1,...,C\}$}
\STATE Assign $\Omega:=\Omega+(\rho_i(\mathbf{x}(t))-\rho_{c_j;i})^2/\rho_{c_j;i}$ 
\ENDFOR
\STATE Assign $L_t[m]:=\exp(-n \cdot (\rho[c_j][t]+\lambda\Omega/C))$
\STATE Assign $s:=s+L_t[m]$
\ENDFOR
\FOR{each top class index $m \in \{1,...,M\}$}
\STATE Assign $L[m]:=L[m]+L_t[m]/s$
\ENDFOR
\ENDFOR
\RETURN class label $c_{m^*}$, where $m^*=\underset{m \in \{1,...,M\}}\argmax L[m]$
\end{algorithmic}
\end{algorithm}

Our complete procedure is presented in Algorithm~\ref{algorithm1}. Its runtime complexity is equal to $O(T(R+MC))$. Here the decision is made only after observation of all $T$ frames. However, it is not difficult to implement online recognition in our approach by looking for the class label $c^*$ right after the next video frame with the same subject is available.

\section{Experimental results}
\label{sec:6}

In most experiments of still-to-video face recognition methods it is usually assumed that the still photos are of high quality and resolution, in frontal view, with normal lighting and neutral expression~\cite{huang2012benchmarking,zhu2015still}. However, in this paper we decided to consider the most challenging experimental setup, in which the training set contains photos gathered in uncontrolled environment~\cite{learned2016labeled}. Namely, we choose all photos and videos of $C=1589$ classes as the intersection of people from the well-known LFW and YTF datasets. It is important to mention that though the YTF dataset contains the videos of celebrities from the LFW, the quality of the images in these datasets are completely different. The training set is filled by all $R=4732$ photos of these $C=1589$ identities taken from the LFW dataset~\cite{learned2016labeled}. Our testing set contains 3353 videoclips from the YTF~\cite{wolf2011face} (in average, 183 frames per person). In our experiments we discovered that it is not necessary to process all frames in each video, hence, only each fifth frame in every video is presented in the testing set, given $T=36$ frames per video in average.

The Caffe framework is applied to extract deep bottleneck features using two publicly available CNN models for unconstrained face recognition, namely, the well-known VGG-13 Network (VGGNet)~\cite{parkhi2015deep}, and the Lightened CNN~\cite{wu2016light}. The fc8 layer of the VGGNet extracts $D=4096$ non-negative features from 224x224 RGB image. The Lightened CNN (version C) extracts $D=256$ (possibly negative) features from 128x128 grayscale image. The outputs of these CNNs were $L_2$ normalized to form the final feature vectors, which are matched using Euclidean distance. The features in the VGGNet are positive, hence we also perform $L_1$ normalization to treat them as the probability distributions and match them using the KL divergence.

\begin{figure}[h!]
\begin{center}
 \includegraphics[width=0.9\linewidth]{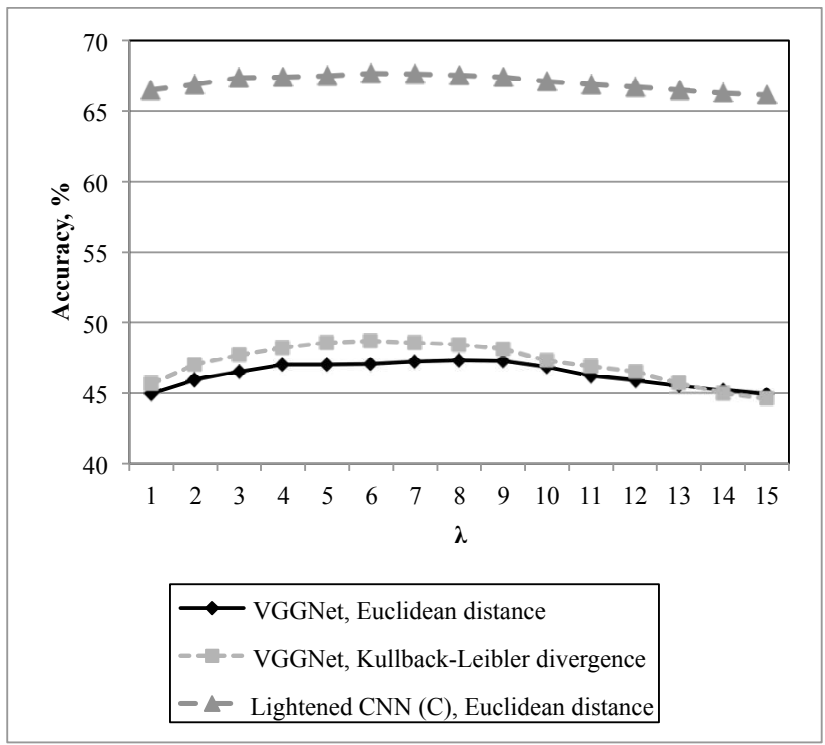}
\end{center}
 \caption{Dependence of the recognition accuracy (\%) on the importance $\lambda$ of regularization term (23), $M=C$.}
\label{fig:1}
\end{figure}

\begin{figure}[h!]
\begin{center}
 \includegraphics[width=0.9\linewidth]{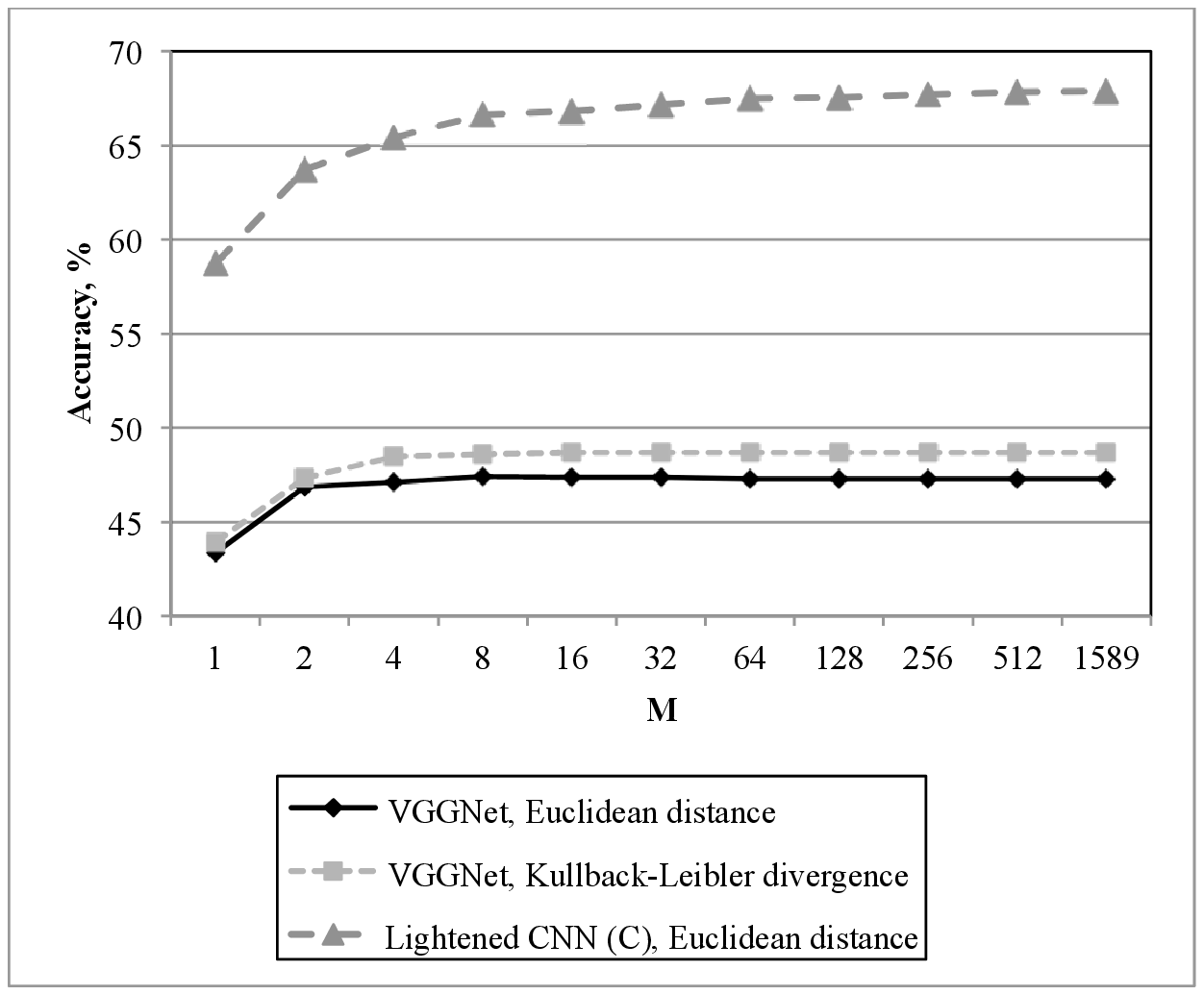}
\end{center}
 \caption{Dependence of the recognition accuracy (\%) on the number of candidate classes $M$, $\lambda=7$.}
\label{fig:2}
\end{figure}

The proposed Algorithm~\ref{algorithm1} is implemented in a stand-alone C++ application with Qt 5 framework. In the first experiments we examined the dependence of the top-1 accuracy on the most important parameters of our algorithm, namely, the importance $\lambda$ of regularization term $\sum\limits_{t=1}^{T} {\sum\limits_{i=1}^{C} {(\rho_i(\mathbf{x}(t))-\rho_{c;i})^2/\rho_{c;i}}}$ (23), and the number of candidate classes $M$. The results are presented in Fig.~\ref{fig:1} and Fig.~\ref{fig:2}, respectively.

Here, firstly, error rates drastically depend on the deep CNN used for feature extraction. The accuracy of the Lightened CNN~\cite{wu2016light} is approximately 20\% higher than the accuracy of much widely used VGGNet~\cite{parkhi2015deep}. It is worth noting that treating the features from the latter network as probability distributions and matching them with the KL divergence (4) allows decreasing error rate at 1.4\% in the best case. Secondly, the U-curves in Fig.~\ref{fig:1} prove that the proper choice of the parameter $\lambda$ can significantly influence the recognition accuracy. It is especially true for the VGGNet features, in which the difference in error rates achieves 2.5\%. However, the optimal value of this parameter is practically identical ($\lambda=6..8$) for all three combination of the CNN and dissimilarity measure. At the same time, the curves in Fig.~\ref{fig:2} reaches stability very fast with an increase of the number of candidate classes $M$. For example, the accuracy for the VGGNet features does not increase further if $M \ge 16$. Though the examination of all $M=C$ classes causes the lowest error rates, this value is not recommended for practical usage due to the very low recognition speed. Hence, in the further part of this paper we use near-optimal values of these parameters: $\lambda=7$ and $M=64$.

\begin{figure}[h!]
\begin{center}
 \includegraphics[width=0.9\linewidth]{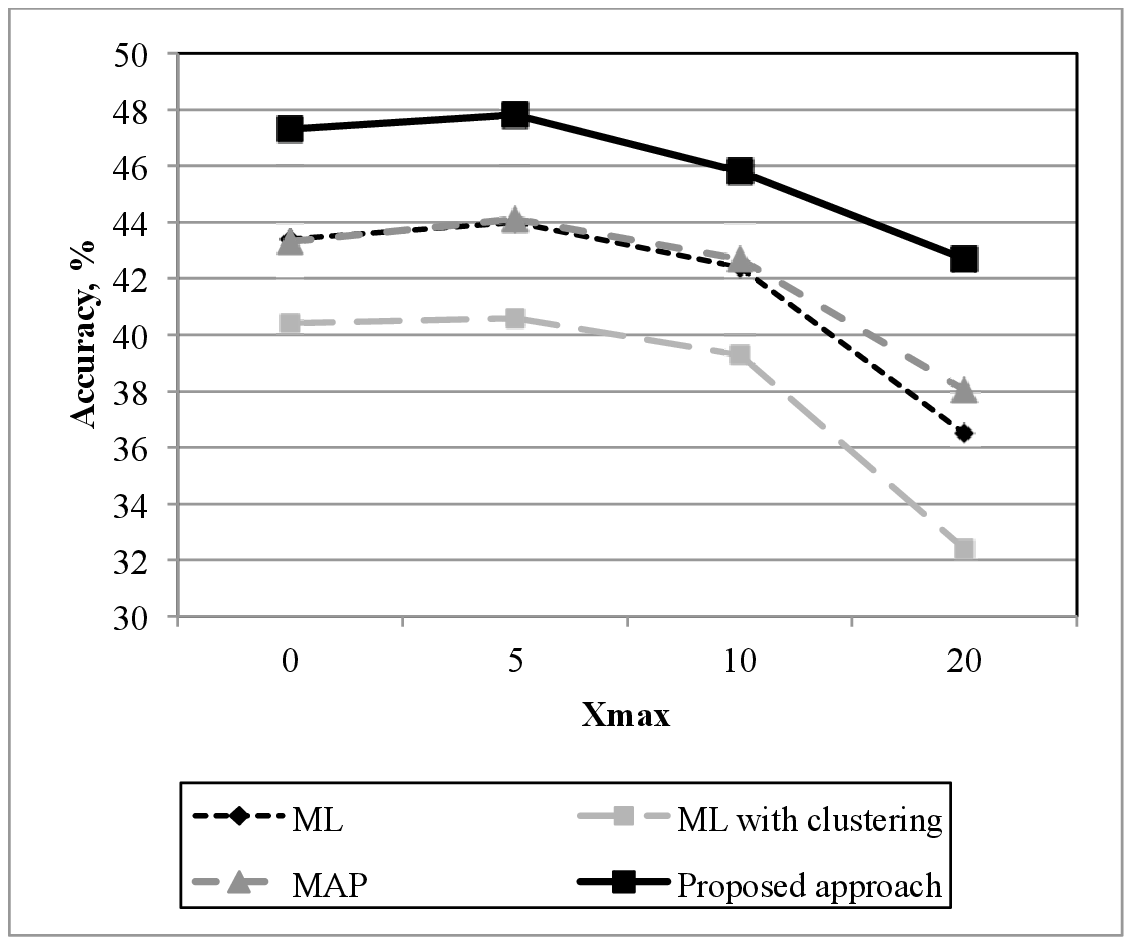}
\end{center}
 \caption{Dependence of the recognition accuracy (\%) on the noise level $X_{\max}$, VGGNet, Euclidean distance.}
\label{fig:3}
\end{figure}

\begin{figure}[h!]
\begin{center}
 \includegraphics[width=0.9\linewidth]{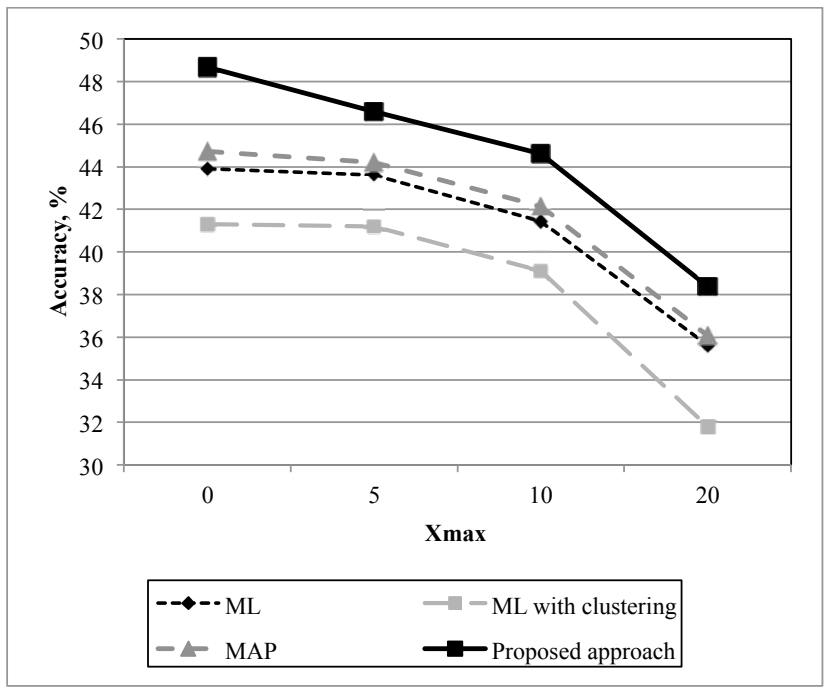}
\end{center}
 \caption{Dependence of the recognition accuracy (\%) on the noise level $X_{\max}$, VGGNet, Kullback-Leibler divergence.}
\label{fig:4}
\end{figure}

\begin{figure}[h!]
\begin{center}
 \includegraphics[width=0.9\linewidth]{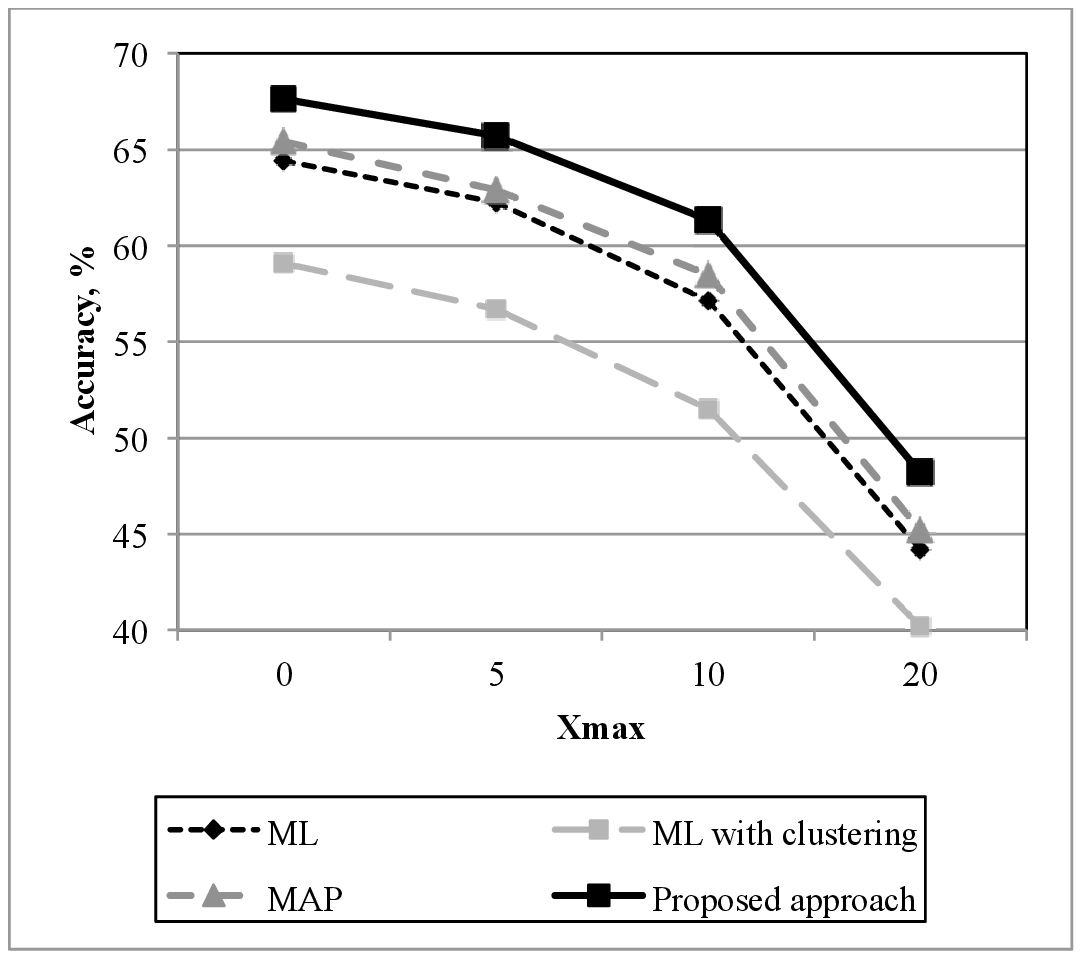}
\end{center}
 \caption{Dependence of the recognition accuracy (\%) on the noise level $X_{\max}$, Lightened CNN (version C), Euclidean distance.}
\label{fig:5}
\end{figure}

In the next experiments we examine the influence of additive noise on the face recognition quality. The uniform random number from the range $[-X_{\max};X_{\max}]$ was added to every pixel of each frame of all testing videoclips, where $X_{\max} \ge 0$ determines the noise level. Original videos from the YTF dataset are recognized, when $X_{\max}=0$. The proposed approach (23) is compared with conventional video face recognition techniques, namely, 1) the ML rule (17) applied to all frames~\cite{savchenko2012adaptive,shakhnarovich2002face}; 2) selection of one representative frame with the k-medoids clustering method ($k=1$) and the NN matching (17) of this frame~\cite{chen2015end,li2014eigen} (hereinafter "ML with clustering"); and 3) estimation of the MAP class for each frame (18) (hereinafter "MAP").
The parameter $n$ of the proposed criterion and MAP rule is tuned in order to provide the highest validation accuracy. The main results of these experiments are shown in Fig.~\ref{fig:3}, Fig.~\ref{fig:4} and Fig.~\ref{fig:5} for VGGNet features matched with the Euclidean distance and the KL divergence, together with the Lightened CNN features, respectively.

Based on these results one can draw the following conclusions. Firstly, though the additive noise leads to degraded recognition accuracy in most cases, the increase of the error rate is rather low even if up to $X_{\max}=10$ number is added to the value of each pixel. Secondly, though the KL divergence outperforms the conventional Euclidean distance for original YTF dataset ($X_{\max}=0$), the latter dissimilarity is more robust to the presence of noise. In fact, it is better to apply more robust probabilistic dissimilarities, which are based on testing for statistical homogeneity of the feature vectors~\cite{savchenko2016search,savchenko2015statistical}. Nevertheless, the Lightened CNN again significantly outperforms the VGGNet even for very high noise levels. Thirdly, though the rather popular accumulation of all frames in one centroid (video face representation~\cite{chen2015end,li2014eigen,yang2016neural}) in the "ML with clustering" method makes it possible to increase performance in $T$ times, its error rate is 2-5\% and 2-6\% higher than the error rate of the baseline ML (17) and the average posterior probability pooling (18)~\cite{kittler1998combining,savchenko2012adaptive}. Moreover, if this approach is applied in online video recognition right after new $t$th frame is observed, the gain in performance over other methods is not so obvious. Really, it is easy to implement criteria (17), (23) in adaptive mode~\cite{savchenko2012adaptive} so that the processing of the $t$th frame is aggregated with the results obtained for recognition of previous $(t-1)$ frames. It is not surprising that the frame selection methods~\cite{huang2013coupling} have more potential in video-to-video recognition or video verification tasks~\cite{yang2016neural}, in which the matching of all pairs of frames from two video clips is very expensive. Finally, the most important conclusion here is the highest recognition accuracy of the proposed algorithm~\ref{algorithm1} in all cases. Our approach is 3-4\% more accurate when compared to other methods. Moreover, though our idea is based on looking for correspondence of the distances $\rho_i(\mathbf{x}(t))$ (17) and $\rho_{c;i}$ (18), (19), the gain in error rate does not significantly degrade even with the presence of noise in the testing videos. It is important to highlight that though the original version of the proposed approach (14) was based on the properties of the KL divergence between positive features, our algorithm can be successfully used with the state-of-the-art distances and arbitrary feature vectors.

In the last part of this section we provide preliminary results for IJB-A dataset~\cite{klare2015pushing}. The IJB-A 1:N protocol is primarily used for testing of the video-to-video recognition. Moreover, each split is used to recognize only 33\% subjects, the rest part can be used to train an algorithm under the strict condition that no such imagery contain the same subjects that are in the test split. Hence, we decided to implement more difficult scenario of the still-to-video recognition with all $C=500$ subjects. We put all $R=5712$ still photos of these subjects to the training set. The testing set contains all 2043 videos from this dataset with approximately 8 frames per video. The average accuracies for all methods described above are shown in Table~\ref{table:1}.

\begin{table}
\begin{center}
\begin{tabular}{|l|ccc|}
\hline
Method & \parbox{2.9cm}{VGGNet, Euclidean distance} & \parbox{2cm}{VGGNet, KL divergence} & \parbox{2.8cm}{Lightened CNN, Euclidean distance} \\
\hline\hline
ML & 68.3 & 67.5 & 42.2\\
ML with clustering & 67.4 & 66.9 & 41.7 \\
MAP & 71.7 & 69.2 & 46.5\\
Proposed approach & 72.8 & 71.6 & 47.8\\
\hline
\end{tabular}
\end{center}
\caption{Recognition accuracy (\%) for IJB-A dataset.}
\label{table:1}
\end{table}

In contrast to the previous experiments, here the features from the VGGNet~\cite{parkhi2015deep} allows classifying videos 25\% more accurately when compared to the Lightened CNN~\cite{wu2016light}, which can be probably caused by the presence of many non-frontal photos in this dataset. However, all the previous conclusions about the still-to-video recognition methods remain the same. The selection of the most representative frame in the ML with clustering is 0.6-0.9\% less accurate that the matching of all frames in criterion (17). The pooling of posterior probabilities (18) here is much preferable than the ML rule (17). At the same time, the error rate of the proposed algorithm~\ref{algorithm1} is 4.5-5.5\% and 1.1-2.5\% lower, than the error rates of the widely-used criteria (17) and (18), respectively.

\section{Conclusion}
\label{sec:7}

In this paper we proposed the novel statistical approach to still-to-video face recognition (Algorithm~\ref{algorithm1}), in which the joint density of distances to all reference images is maximized (7), (8). We have shown that this approach is implemented by introduction of a special regularization term (23) in the nearest neighbor matching (17) of high-dimensional features from the outputs of the deep CNNs. In this regularization it is assumed that the resulted class $c^*(t)$ for recognition of individual $t$th video frame is reliable only if the distances between the features of this frame and the $r$th still image are approximately equal to the distances between the reference images from the $c^*(t)$ and $c(r)$ classes for all $r \in \{1,...,R\}$. This assumption is known to be asymptotically correct for the KL divergence and rather simple probabilistic model from Section~\ref{sec:3} ~\cite{kullback1997information}. However, it was experimentally demonstrated that the proposed approach can be combined even with the state-of-the-art Euclidean distance (Fig.~\ref{fig:4}). Moreover, our algorithm allows to increase accuracy even for general feature vectors with possible negative values (Fig.~\ref{fig:5}). We demonstrated how to tune the parameters of the proposed algorithm (Fig.~\ref{fig:2}) in order to drastically reduce its computation complexity by proper choice of $M$ candidate classes (compare the criterion (23) with the original one (19)).

The main direction for further research of the proposed algorithm is its applications with more accurate approximation of the distance probability distributions, e.g., the usage of the more appropriate Weibull distribution~\cite{burghouts2008distribution}. Secondly, it is important to make our regularization smoother by taking into account the temporal coherence of sequential frames~\cite{liu2003video}. Thirdly, it is necessary to combine our approach with the modern techniques with frames weighting in the aggregation module (22)~\cite{yang2016neural} to make distinction between frames with different quality ~\cite{huang2013coupling}. 
Finally, though the deep bottleneck features are known to be highly embedded in the manifold space, non-Euclidean metrics, e.g, such techniques as the point-to-set metric learning~\cite{huang2014learning} or geometry aware feature matching~\cite{harandi2014manifold} can also be discussed in the future experiments.

\section{Acknowledgement}
The paper is supported by Russian Federation President grant no. MD-306.2017.9. The work in Section 3 was conducted by A.V. Savchenko at Laboratory of Algorithms and Technologies for Network Analysis, National Research University Higher School of Economics and supported by RSF grant 14-41-00039.

\end{document}